\documentclass[letterpaper]{article} 
\usepackage{aaai23}  
\usepackage{times}  
\usepackage{helvet}  
\usepackage{courier}  
\usepackage[hyphens]{url}  
\usepackage{graphicx} 
\usepackage{natbib}  
\usepackage{caption} 
\usepackage{amsfonts}
\usepackage{algorithm}
\usepackage{algorithmic}

\usepackage{newfloat}
\usepackage{listings}
\DeclareCaptionStyle{ruled}{labelfont=normalfont,labelsep=colon,strut=off} 
\floatstyle{ruled}
\newfloat{listing}{tb}{lst}{}
\floatname{listing}{Listing}
\title{Privacy-Preserving Representation Learning for Text-Attributed Networks with Simplicial Complexes}
\author {
    Huixin Zhan\textsuperscript{\textdagger} and Victor S. Sheng\textsuperscript{\textasteriskcentered}
}
\affiliations{
Department of Computer Science, Texas Tech University, Lubbock, TX 79409-3104, USA\\\{\textsuperscript{\textdagger}Huixin.Zhan, \textsuperscript{\textasteriskcentered}Victor.Sheng\}@ttu.edu
}

\usepackage{bibentry}

\begin{document}

\maketitle

\begin{abstract}
Although recent network representation learning (NRL) works in text-attributed networks demonstrated superior performance for various graph inference tasks, learning network representations could always raise privacy concerns when nodes represent people or human-related variables. Moreover, standard NRLs that leverage structural information from a graph proceed by first encoding pairwise relationships into learned representations and then analysing its properties. This approach is fundamentally misaligned with problems where the relationships involve multiple points, and topological structure must be encoded beyond pairwise interactions. Fortunately, the machinery of topological data analysis (TDA) and, in particular, simplicial neural networks (SNNs) offer a mathematically rigorous framework to learn higher-order interactions between nodes. It is critical to investigate if the representation outputs from SNNs are more vulnerable compared to regular representation outputs from graph neural networks (GNNs) via pairwise interactions. In my dissertation, I will first study learning the representations with text attributes for simplicial complexes (RT4SC) via SNNs. Then, I will conduct research on two potential attacks on the representation outputs from SNNs: (1) membership inference attack, which infers whether a certain node of a graph is inside the training data of the GNN model; and (2) graph reconstruction attacks, which infer the confidential edges of a text-attributed network. Finally, I will study a privacy-preserving deterministic differentially private alternating direction method of multiplier to learn secure representation outputs from SNNs that capture multi-scale relationships and facilitate the passage from local structure to global invariant features on text-attributed networks. 
  
\end{abstract}

\section{Research Problems}
In text attributed networks, nodes are characterized by textual information. For example, in social networks, the nodes (users) are
linked with other nodes by friend relationships and each node is associated with text like their profile information. Recent works show that leveraging both textual information and topological structure into network representation learning (NRL) benefits downstream tasks performances~\citep{chen2021effective,zhang2019attributed}.  However, most real-world graphs associated with people or human-related activities are often sensitive and might contain confidential information~\citep{sajadmanesh2021locally}. Releasing the representations of nodes in real world graphs gives adversaries a potential way to infer the sensitive information of nodes and edges. Thus, it is critical to develop privacy-preserving representations for applications using graphs that require users' sensitive data to fulfill users' privacy requirements.

\textbf{Challenges}
\textit{Current NRL works have a few remaining problems. First, training a Graph Neural Network (GNN) from private node data is a challenging task due to the absence of a trusted third party. Second, typical NRLs, e.g., diffusion mechanisms and random walks on graphs, only assess the pairwise relationships between two individual nodes, but ignore the higher-order interactions between nodes. Third, current GNNs ignore important interactions between global invariant features, i.e., data characteristics which are invariant under continuous transformations such as stretching, bending, and compressing.} Recently, simplicial neural networks (SNNs) offer a mathematically rigorous framework to evaluate not only higher-order interactions, but also global invariant features of the observed graph to systematically learn topological structures. These features occur in the form of homological features, intuitively perceived as holes, or voids, in any desired dimension. It is important to investigate -$\mathbf{Q1}$: how to build representation outputs from SNNs (RO\_SNNs) that integrate attribute textual information with multi-scale relationships between nodes and -$\mathbf{Q2}$: to study if the RO\_SNNs are exposed to new threats and if they are more vulnerable compared to regular representation outputs from typical GNNs. Besides, -$\mathbf{Q3}$: it is also challenging to propose \textit{secure RO\_SNNs} that even if an adversary has access to user representation outputs in the database, that adversary will still be unable to learn too much about the user's sensitive data.  
\section{Research Plan}
Currently, there are a few SNNs~\citep{keros2022dist2cycle,chen2021topological} that utilize GNN models for learning functions parametrized by the homological features to learn topological relationships of the underlying simplicial complexes. All existing SNNs are not focusing on NRL on text-attributed networks. It is non-trivial to incorporate text representations in existing SNNs due to SNNs cannot easily handle additional information during its graph convolution. Fortunately, according to the theoretical finding that NRLs are equivalent to factorize an affinity matrix $M$ derived from the adjacency matrix of the original network~\citep{yang2015network}. In $\mathbf{Q1}$, we first develop the RT4SC. The learning process includes two stages. In the first stage, we integrate text representations with regular pairwise node interactions via factorizing $M$ into the product of three matrices as $M=W^THT$, where $W \in \mathbb{R}^{k\times \cdot }$, $H \in \mathbb{R}^{k \times t}$ and
text features $T \in \mathbb{R}^{t \times \cdot}$. Then we can concatenate $W$ and $HT$ as $2k$-dimensional representations of nodes. In the second stage, we will enrich the node representations by: 1) first extracting local topological side information from subgraphs using persistent homology of the small neighborhoods of nodes; and then 2) incorporating the extracted local topological side information into the local GNN algorithm for NRL. In $\mathbf{Q2}$, we will first investigate the membership inference attack and show that an adversary can distinguish which node participates in the training of the GNN via training an inference model to recognize differences between the prediction of the model trained with the record and that of the model trained without the record. Second, the server can aggregate nodes’ representations with their neighbors to learn better user representations for improving its services. This means if there is an edge between two nodes, then their RO\_SNNs should be closer. Therefore, a potential adversary could possibly recover the sensitive edge information (e.g., friend lists) via a machine learning classifier that simply measures distance differences of the RO\_SNNs. Thus, we will study whether representations can be inverted to recover the graph used to generate
them. Graph reconstruction attacks (GRAs) try to infer the edges of graphs. Regular GRAs predict edges via measuring distance differences between linked node pairs and unlinked node pairs via clustering~\citep{he2021stealing}. However, we will propose a GRA that utilizes a graph-decoder to minimize the reconstruction loss of the generated adjacency matrix via backpropagation. We will further perform both attacks on outputs from regular GNNs and measure if RO\_SNNs are more vulnerable compared to regular outputs from GNNs. In $\mathbf{Q3}$, I will study a privacy-preserving deterministic differentially private alternating direction method of multiplier, i.e., D$^2$-ADMM, to learn secure RO\_SNNs that not only capture multi-scale relationships, but also could defend the potential attacks on an untrusted server.
\paragraph{Completed Research and Timeline}
Me and my Ph.D. advisor have been researching privacy-preserving text representations. In our paper~\citep{zhan2021multi}, we show that some of the hidden private information correlates with the output labels and therefore can be learned by a neural network. In such a case, there is a tradeoff between the utility of the representation and its privacy. We explicitly cast this problem as multi-objective optimization and propose a multiple-gradient descent algorithm that enables the efficient application of the Frank-Wolfe algorithm to search for the optimal utility privacy configuration of the text classification network. Our prior work~\citep{zhan2022new} also show that it is challenging to protect privacy while preserving important semantic information about an input text.  In particular, the threats are (1) these representations reveal sensitive attributes, no matter if they explicitly exist in the input text and (2) the representations can be partially recovered via generative models. In our recent paper~\citep{zhan2023graph}, we propose a GRA to recover a graph’s adjacency matrix from three types of representation outputs, i.e., representation outputs from graph convolutional networks, graph attention networks, and SNNs. We find that SNN outputs obtain the highest precision and AUC on five real-world networks. Therefore, the SNN outputs reveal the lowest privacy-preserving ability to defend the GRAs. Thus, it calls for future research towards building more private and higher-order representations that could defend the potential threats. My research timeline is as follows: By the date of submission, I addressed the first part of $\mathbf{Q2}$. By the workshop date, I plan to complete $\mathbf{Q1}$ and $\mathbf{Q2}$ and I will complete $\mathbf{Q3}$ after the workshop. 

\bibliography{aaai23}

\begin{thebibliography}{10}
\providecommand{\natexlab}[1]{#1}

\bibitem[{Chen et~al.(2021)Chen, Zhong, Li, Wang, Qian, and
  Tu}]{chen2021effective}
Chen, J.; Zhong, M.; Li, J.; Wang, D.; Qian, T.; and Tu, H. 2021.
\newblock Effective deep attributed network representation learning with
  topology adapted smoothing.
\newblock \emph{IEEE Transactions on Cybernetics}.

\bibitem[{Chen, Coskunuzer, and Gel(2021)}]{chen2021topological}
Chen, Y.; Coskunuzer, B.; and Gel, Y. 2021.
\newblock Topological relational learning on graphs.
\newblock \emph{Advances in Neural Information Processing Systems}, 34:
  27029--27042.

\bibitem[{He et~al.(2021)He, Jia, Backes, Gong, and Zhang}]{he2021stealing}
He, X.; Jia, J.; Backes, M.; Gong, N.~Z.; and Zhang, Y. 2021.
\newblock Stealing links from graph neural networks.
\newblock In \emph{30th USENIX Security Symposium}, 2669--2686.

\bibitem[{Keros, Nanda, and Subr(2022)}]{keros2022dist2cycle}
Keros, A.~D.; Nanda, V.; and Subr, K. 2022.
\newblock Dist2cycle: A simplicial neural network for homology localization.
\newblock In \emph{Proceedings of the 36th AAAI Conference on Artificial
  Intelligence}, volume~36, 7133--7142.

\bibitem[{Sajadmanesh and Gatica-Perez(2021)}]{sajadmanesh2021locally}
Sajadmanesh, S.; and Gatica-Perez, D. 2021.
\newblock Locally private graph neural networks.
\newblock In \emph{Proceedings of the 2021 ACM SIGSAC Conference on Computer
  and Communications Security}, 2130--2145.

\bibitem[{Yang et~al.(2015)Yang, Liu, Zhao, Sun, and Chang}]{yang2015network}
Yang, C.; Liu, Z.; Zhao, D.; Sun, M.; and Chang, E. 2015.
\newblock Network representation learning with rich text information.
\newblock In \emph{Twenty-fourth international joint conference on artificial
  intelligence}.

\bibitem[{Zhan et~al.(2021)Zhan, Zhang, Hu, and Sheng}]{zhan2021multi}
Zhan, H.; Zhang, K.; Hu, C.; and Sheng, V. 2021.
\newblock Multi-objective Privacy-preserving Text Representation Learning.
\newblock In \emph{Proceedings of the 30th ACM International Conference on
  Information \& Knowledge Management}, 3612--3616.

\bibitem[{Zhan et~al.(2022)Zhan, Zhang, Hu, and Sheng}]{zhan2022new}
Zhan, H.; Zhang, K.; Hu, C.; and Sheng, V. 2022.
\newblock New Threats to Privacy-preserving Text Representations.
\newblock In \emph{Proceedings of the 55th Hawaii International Conference on
  System Sciences}, 768--777.

\bibitem[{Zhan et~al.(2023)Zhan, Zhang, Lu, and Sheng}]{zhan2023graph}
Zhan, H.; Zhang, K.; Lu, K.; and Sheng, V. 2023.
\newblock Measuring the Privacy Leakage via Graph Reconstruction Attacks on
  Simplicial Neural Networks (Student Abstract).
\newblock In \emph{Proceedings of the 37th AAAI Conference on Artificial
  Intelligence}.

\bibitem[{Zhang et~al.(2019)Zhang, Yin, Zhu, and Zhang}]{zhang2019attributed}
Zhang, D.; Yin, J.; Zhu, X.; and Zhang, C. 2019.
\newblock Attributed network embedding via subspace discovery.
\newblock \emph{Data Mining and Knowledge Discovery}, 33(6): 1953--1980.

\end{thebibliography}

\end{document}